\documentclass[sigconf]{acmart}
\AtBeginDocument{%
  }

\copyrightyear{2026}
\acmYear{2026}
\setcopyright{acmlicensed}
\acmConference[arXiv Preprint]{arXiv Preprint}{2026}{arXiv.org}
\acmBooktitle{arXiv Preprint (arXiv Preprint), 2026, arXiv.org}
\acmDOI{10.48550/arXiv.2601.20257}
\acmISBN{978-1-4503-XXXX-X/26/01}

\usepackage{bbm}
\usepackage{multirow}
\usepackage{algorithm,algorithmic}
\usepackage{enumitem}
\usepackage{color,xspace}
\usepackage{xcolor}
\usepackage{soul,ulem}
\usepackage{amsmath}
\usepackage{bbm}
\usepackage{colortbl}
\usepackage{subcaption}
\usepackage{graphicx}
\usepackage{textcomp}
\usepackage{booktabs}
\usepackage{makecell}

\usepackage{tcolorbox}  
\tcbuselibrary{breakable}   

\begin{document}

\title{C2:Cross learning module enhanced decision transformer with Constraint-aware loss for auto-bidding}

\author{Jinren Ding}
\affiliation{%
  \institution{Kuaishou Technology}
  \city{Beijing}
  \country{China}}
\email{dingjinren@kuaishou.com}

\author{Xuejian Xu}
\affiliation{%
  \institution{Kuaishou Technology}
  \city{Beijing}
  \country{China}}
\email{xuxuejian@kuaishou.com}

\author{Shen Jiang}
\authornote{Corresponding authors.}
\affiliation{%
  \institution{Kuaishou Technology}
  \city{Beijing}
  \country{China}}
\email{jiangshen@kuaishou.com}

\author{Zhitong Hao}
\affiliation{%
  \institution{Kuaishou Technology}
  \city{Beijing}
  \country{China}}
\email{haozhitong@kuaishou.com}

\author{Jinhui Yang}
\affiliation{%
  \institution{Kuaishou Technology}
  \city{Beijing}
  \country{China}}
\email{yangjinhui@kuaishou.com}

\author{Peng Jiang}
\affiliation{%
  \institution{Kuaishou Technology}
  \city{Beijing}
  \country{China}}
\email{jiangpeng@kuaishou.com}

\renewcommand{\shortauthors}{Ding et al.}

\begin{abstract}

Decision Transformer (DT) shows promise for generative auto-bidding by capturing temporal dependencies, but suffers from two critical limitations: insufficient cross-correlation modeling among state, action, and return-to-go (RTG) sequences, and indiscriminate learning of optimal/suboptimal behaviors. To address these, we propose C2, a novel framework enhancing DT with two core innovations: (1) a Cross Learning Block (CLB) via cross-attention to strengthen inter-sequence correlation modeling; (2) a Constraint-aware Loss (CL) incorporating budget and Cost-Per-Acquisition (CPA) constraints for selective learning of optimal trajectories. Extensive offline evaluations on the AuctionNet dataset demonstrate consistent performance gains (up to 3.23\% over state-of-the-art method) across diverse budget settings; ablation studies verify the complementary synergy of CLB and CL, confirming C2's superiority in auto-bidding. The code for reproducing our results is available at: \url{https://github.com/Dingjinren/C2}.

\end{abstract}




\maketitle

\section{Introduction}
Auto-bidding has become an indispensable component of modern online advertising ecosystems~\cite{zhao2018deep, zhao2020jointly,gao2024smlp4rec,gao2024hierrec}, serving as a critical bridge between advertisers' diverse objectives (e.g., maximizing conversions, controlling cost-per-action) and dynamic market environments. With the exponential growth of ad impressions and the increasing complexity of user behaviors~\cite{aggarwal2024auto}, traditional manual bid management and static rule-based strategies~\cite{chen2011real,yu2017online} have become impractical, as they fail to adapt to real-time fluctuations in market competition and advertisers' evolving preferences~\cite{jha2024optimizing,liu2024sequential}. This has driven the shift toward data-driven automated solutions, among which reinforcement learning (RL)-based methods have been widely explored by framing the bidding problem as a Markov Decision Process (MDP)~\cite{ye2019deep,zhao2019deep}.

However, RL-based approaches~\cite{ye2019deep, cai2017real,gao2023autotransfer,zhao2021dear} inherently suffer from a critical limitation: the MDP framework's assumption of state independence disregards the long-range temporal dependencies embedded in bidding sequences. In real-world auto-bidding scenarios, subsequent bidding outcomes are strongly correlated with historical trajectories (e.g., past bid amounts, user engagement patterns, and remaining budget dynamics), making the MDP-based state abstraction overly simplistic and leading to suboptimal decision-making. To tackle this issue, the dominant generative solution lies in Decision Transformer (DT), a sequence-modeling framework tailored to capture complex temporal correlations and contextual information inherent in bidding processes.

Specifically, Decision Transformer (DT)~\cite{chen2021decision} has established itself as a powerful framework in generative auto-bidding, owing to its dual strengths: it effectively captures intricate temporal dependencies in bidding sequences, and seamlessly integrates multi-dimensional inputs (states, actions, and return-to-go rewards) to generate optimal next-step bidding actions, which perfectly match the adaptive decision-making requirements of dynamic advertising scenarios. Moreover, its advantages in modeling long-range dependencies and fusing multi-factor information address the limitations of traditional isolated-factor bidding models, opening up a promising avenue for advancing bidding strategy performance in practical advertising ecosystems.

However, several issues have emerged when applying DT-based models~\cite{gao2025generative,li2024gas,li2025ebaret} to real-world auto-bidding scenarios. First, when processing sequential information, conventional DT models simply stack the embeddings of states, actions, and return-to-go values before feeding them into the Transformer, which results in insufficient modeling of the correlations among these three components—especially when facing complex advertising objectives involving interdependent parameters like cost thresholds. Second, the intrinsic nature of DT models lies in mimicking existing historical bidding action sequences, which means they indiscriminately learn both optimal and suboptimal bidding behaviors without differentiation, restricting themselves to documented behavioral patterns and even suffering from behavioral collapse.

To tackle these challenges, we propose a novel approach: \textbf{C}ross learning block-enhanced decision transformer with \textbf{C}onstraint-aware loss for auto-bidding (C2). First, to tackle the insufficient modeling of correlations among the input sequences of states, actions, and return-to-go values in conventional DT models, we introduce the Cross Learning Block-enhanced DT (CLB-DT), which enables more in-depth learning of the correlational information among these three components through cross-attention~\cite{brown2020language, vaswani2017attention, melnychuk2022causal}. Second, to resolve the issue that DT models indiscriminately learn all historical bidding action sequences, we adopt a Constraint-aware Loss (CL) to guide the targeted learning of bidding sequences—specifically, promoting the learning of optimal bidding behaviors while suppressing suboptimal ones~\cite{edelman2007internet,aggarwal2009general}.

Our contributions are summarized as follows:  

\begin{itemize}[leftmargin=*]  
    \item We identify critical limitations of conventional DT in real-world auto-bidding, namely inadequate correlation modeling among state-action-return-to-go sequences and indiscriminate learning of historical behaviors, highlighting the need for targeted improvements.
    \item We propose a novel C2 framework, integrating two core innovations: (1) A Cross Learning Block (CLB) with cross-attention that replaces traditional Transformer blocks to enhance correlation modeling of multi-dimensional sequential inputs; (2) A Constraint-aware Loss that guides the model to prioritize optimal bidding behaviors by incorporating budget and other Key Performance Indicator (KPI) constraints.
    \item We validate our approach through comprehensive offline experiments. Evaluations on the AuctionNet dataset~\cite{su2024a} show C2 outperforms state-of-the-art baselines (up to 3.23\% gain over GAVE) across 50\%-150\% budget settings; ablation studies confirm CLB and CL contribute 10.2\% and 7.2\% gains respectively, with 15.3\% synergetic improvement over vanilla DT, verifying C2's effectiveness.
\end{itemize}
\section{Preliminary}
This section first formalizes the auto-bidding problem with core constraints in real-world advertising scenarios, then introduces the fundamental DT-based sequential decision-making method, laying the foundation for subsequent improvements.

\subsection{Auto-Bidding Problem}\label{sec:autobidding_prob}

Consider a discrete-time auto-bidding process where \( N \) impression opportunities arrive sequentially (\( i = 1, 2, \dots, N \)). An advertiser submits a bid \( b_i \) for each impression to participate in real-time auctions; under the industry-standard generalized second-price (GSP) mechanism~\cite{edelman2007internet,aggarwal2009general}, the advertiser wins if \( b_i \) exceeds the highest competing bid \( b_i^- \). Let \( x_i \in \{0, 1\} \) denote the auction outcome (1 for winning, 0 otherwise), and \( c_i = b_i^- \) (cost of winning in GSP auctions).

The core objective is to maximize total value from won impressions (\( v_i \in \mathbb{R}^+ \), e.g., CTR/CVR-weighted value), with bidding strategies satisfying budget and CPA constraints (Multi-Constrained Bidding setting). The auto-bidding problem is formally formulated as:
\begin{equation}
\begin{aligned}
\max_{\{b_i\}_{i=1}^N} &\quad \sum_{i=1}^N x_i v_i \\
\text{s.t.} \quad &\sum_{i=1}^N x_i c_i \leq B, \\
&\frac{\sum_{i=1}^N x_i c_i}{\sum_{i=1}^N x_i v_i} \leq C,  \\
&x_i = 
\begin{cases} 
1 & \text{if } b_i > b_i^-, \\
0 & \text{otherwise},
\end{cases}
\label{eq:goal}
\end{aligned}
\end{equation}

where \( B \in \mathbb{R}^+ \) is the total campaign budget, and \( C \in \mathbb{R}^+ \) is the maximum allowable CPA threshold. As shown in prior work, the optimal bid \( b_i^* \) can be parameterized as a linear combination of \( v_i \) and constraint-related terms:
\begin{equation}
b_i^* = \lambda_0 v_i + \lambda_1 C v_i,
\label{eq:optimal_bid}
\end{equation}

where \( \lambda_0 \) (regulating value-budget trade-off) and \( \lambda_1 \) (adjusting CPA adherence) balance objectives and constraints. However, real-world dynamics (e.g., fluctuating competition and user behavior) hinder direct parameter computation, so DT models are needed to adaptively learn optimal bidding strategies from historical data.

\subsection{DT-based Method}\label{sec:dt_method}

By framing sequential decision-making as conditional sequence modeling, Decision Transformer (DT)~\cite{chen2021decision} effectively captures long-range temporal dependencies in auto-bidding sequences. Unlike traditional RL methods, DT avoids restrictive MDP assumptions (which neglect trajectory correlation structures) and leverages the Transformer's superior sequence modeling capacity to map historical segments to optimal actions.

For auto-bidding tasks, DT characterizes the decision process via four core components forming a trajectory segment \( \tau = \{\mathbf{R}, \mathbf{S}, \mathbf{A}\} = \left\{ (r_t, s_t, a_t) \right\}_{t-M}^t \), where \( M \) is the input sequence length. Detailed definitions are as follows:

\begin{itemize}[leftmargin=*]
    \item \textbf{State Sequence} \( \mathbf{S} = (s_{t-M}, \dots, s_t) \): Multi-dimensional feature vectors depicting the advertising ecosystem's real-time state at each \( t \), including: (1) remaining budget (\( B_t = B - \sum_{k=1}^{t-1} x_k c_k \)); (2) remaining campaign duration; (3) historical bidding metrics (e.g., hourly average win rate); and (4) real-time cost-rate \( \frac{\sum_{k=1}^{t-1} x_k c_k}{\sum_{k=1}^{t-1} x_k v_k} \) for constraint compliance.

    \item \textbf{Action Sequence} \( \mathbf{A} = (a_{t-M}, \dots, a_t) \): Bidding parameters adjusted at each \( t \). Consistent with Equation~\eqref{eq:optimal_bid}, \( a_t = \lambda_t = \lambda_{0,t} + \lambda_{1,t} C \), the unified parameter determining actual bids \( b_t = \lambda_t v_i \).

    \item \textbf{Reward Sequence} \( \mathbf{RW} = (rw_{t-M}, \dots, rw_t) \): Cumulative value from impressions in \( [t, t+1] \), computed as \( rw_t = \sum_{k \in \mathcal{I}_t} x_k v_k \) ( \( \mathcal{I}_t \) is the set of impressions in this window).

    \item \textbf{Return-to-Go (RTG) Sequence} \( \mathbf{R} = (r_{t-M}, \dots, r_t) \): \( r_t = \sum_{k=t}^T rw_k \), the total remaining reward from \( t \) to campaign end. As a critical conditional signal, RTG guides DT to learn long-term goal-aligned strategies.
\end{itemize}

The standard DT training pipeline has two key steps: concatenate \( r_t, s_t, a_t \) embeddings into a sequential vector, then feed it into a causal Transformer encoder to model inter-component dependencies. The model is optimized to predict \( a_{t+1} \) from the historical segment \( (r_{t-M}, s_{t-M}, a_{t-M}, \dots, r_t, s_t, a_t) \). Despite its strengths, this paradigm has two drawbacks: (1) naive concatenation fails to model cross-correlations between \( r_t, s_t, a_t \) (exacerbated in complex scenarios); (2) lack of constraint-aware objectives leads to indiscriminate learning of optimal/suboptimal behaviors.

DT-based methods are often evaluated on AuctionNet~\cite{su2024a}, which provides large-scale simulated trajectories for fair comparison. Implementations follow prior work~\cite{li2024gas,guo2024generative} using causal Transformers with fixed hyperparameters (e.g., layers, attention heads) optimized via AdamW~\cite{loshchilov2017decoupled}, ensuring consistency with evaluation protocols.
\section{Methodology}

Here, we will detail C2's overview and key components. 

\begin{figure*}[t]
    \centering
    \includegraphics[width=0.87\linewidth]{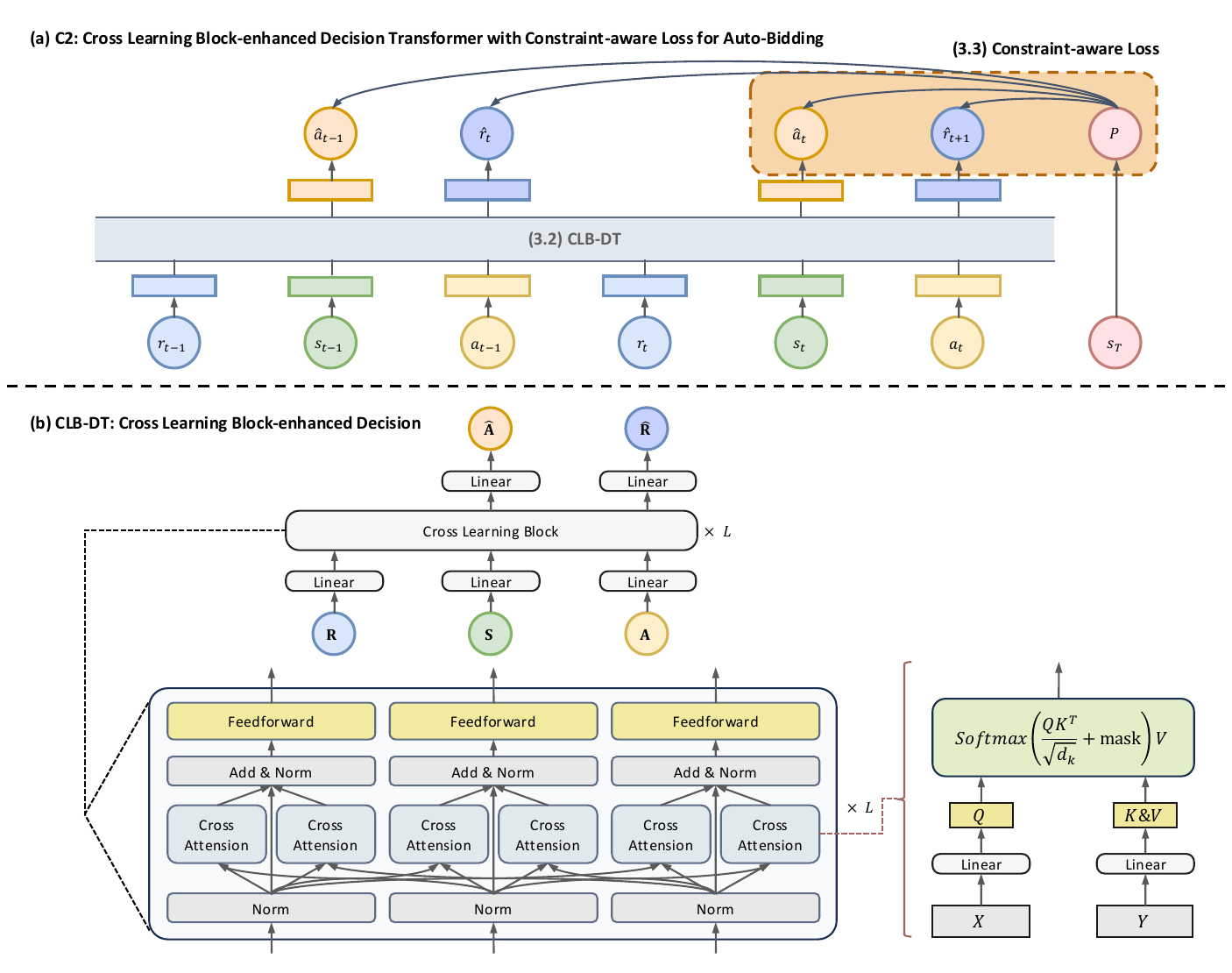}
    \vspace{-5mm}
    \caption{Overall structure of C2.}
    \label{fig:overview}
    \vspace{-5mm}
\end{figure*}

\subsection{C2 Overview}
As illustrated in Figure~\ref{fig:overview}(a), the proposed C2 framework adopts a DT-like architecture, which takes sequences of states, actions, and RTGs as inputs and outputs the optimal subsequent bidding action. Unlike conventional DT models that suffer from two critical limitations—insufficient capacity to capture cross-correlations among state, action, and RTG sequences and indiscriminate learning of all historical bidding samples—C2 introduces two key improvements to address these issues. Specifically, the Cross Learning Block-enhanced DT (CLB-DT) optimizes the model architecture to strengthen the modeling of correlations among the three input sequences, while the Constraint-aware Loss (CL) leverages the constraints of Multi-Constrained Bidding (MCB) to compute adaptive penalty terms. These penalty terms are incorporated into the loss function to guide the model toward selective learning of high-quality bidding trajectories while suppressing suboptimal ones.

\subsection{CLB-DT}\label{sec:opt}
Our Cross Learning Block-enhanced Decision Transformer (CLB-DT) adopts a multi-input unified architecture, which takes three distinct sequences (each of length \(M\)) as inputs: the state sequence \(\mathbf{S}\), action sequence \(\mathbf{A}\), and return-to-go (RTG) sequence \(\mathbf{R}\). Specifically, we introduce the Cross Learning Block (CLB) to replace the vanilla Transformer blocks in the original DT architecture, thereby enabling the model to fully capture the cross-correlations among the three input sequences. The architecture is illustrated in Figure~\ref{fig:overview}(b).

\paragraph{Cross Learning Block.} 
Let \(b = 1, \ldots, B\) index the distinct Cross Learning Blocks. Each block takes three parallel hidden state sequences \(\mathbf{S}^b, \mathbf{A}^b, \mathbf{R}^b\) as inputs (outputs of the \(b\)-th block) and outputs updated sequences \(\mathbf{S}^{b+1}, \mathbf{A}^{b+1}, \mathbf{R}^{b+1}\), where the hidden state dimension is denoted by \(d_h\). For the first block (\(b=1\)), the input sequences are initialized with positional encodings derived from time steps \(\mathbf{T}\), combined with sequence-specific linear encodings:
\begin{align}
    \begin{split}
        & \mathbf{T}^0 = \operatorname{Linear}_T(\mathbf{T}), \quad \mathbf{S}^1 = \operatorname{Linear}_S(\mathbf{S}) + \mathbf{T}^0, \\
        & \mathbf{A}^1 = \operatorname{Linear}_A(\mathbf{A}) + \mathbf{T}^0, \quad \mathbf{R}^1 = \operatorname{Linear}_R(\mathbf{R}) + \mathbf{T}^0,
    \end{split}
    \label{eq:init_encoding}
\end{align}
where \(\operatorname{Linear}_T, \operatorname{Linear}_S, \operatorname{Linear}_A, \operatorname{Linear}_R\) are independent learnable linear layers for positional encoding and sequence-specific encoding, respectively. For blocks with \(b \ge 2\), the inputs \(\mathbf{S}^b, \mathbf{A}^b, \mathbf{R}^b\) are directly the outputs of the preceding block (\(b-1\)).

Each Cross Learning Block integrates three core components: (i) masked cross-attention, (ii) feedforward layer (FF), and (iii) layer normalization (LN), which collaborate to model cross-sequence dependencies and non-linear feature transformations~\cite{radford2019language, brown2020language, vaswani2017attention, melnychuk2022causal}.

\underline{(i) Masked cross-attention} extends standard cross-attention with an attention mask to regulate interactions between two distinct sequences (denoted as \(\mathrm{X}\) and \(\mathrm{Y}\)). Its core consists of three learnable components—Query (\(Q\)), Key (\(K\)), and Value (\(V\))—derived via linear transformations:
\begin{align}
    Q &= Q(\mathrm{X}^b) = \mathrm{X}^b W_Q + \mathbf{1} b_Q^\top, \\
    K &= K(\mathrm{Y}^b) = \mathrm{Y}^b W_K + \mathbf{1} b_K^\top, \\ 
    V &= V(\mathrm{Y}^b) = \mathrm{Y}^b W_V + \mathbf{1} b_V^\top,
    \label{eq:qkv_computation}
\end{align}
where \(W_Q, W_K, W_V \in \mathbb{R}^{d_h \times d_k}\) are learnable weight matrices, \(b_Q, b_K, b_V \in \mathbb{R}^{d_k}\) are bias vectors, \(\mathbf{1}\) is a column vector of ones for bias broadcasting, and \(d_k\) denotes the dimension of Query/Key/Value (typically \(d_k = d_h\)). The superscript \(b\) indicates the block index (consistent with input sequences), and batch dimensions are omitted for brevity.

The attention mask is a binary matrix that suppresses invalid interactions: invalid positions are assigned a large negative value (e.g., \(-10^4\)) to inhibit attention flow, while valid positions are set to zero. The masked cross-attention operation is defined as:
\begin{align}
\operatorname{Attn}(Q, K, V) = \operatorname{softmax}\bigg(\frac{QK^\top}{\sqrt{d_k}} + \text{mask}\bigg) V,
\label{eq:attention}
\end{align}
which ensures the model focuses only on relevant cross-sequence information. The computation proceeds in three steps: (1) compute a similarity matrix via \(QK^\top/\sqrt{d_k}\) to quantify relevance between \(\mathrm{X}\) and \(\mathrm{Y}\); (2) apply the mask to suppress invalid positions; (3) normalize scores via softmax to generate attention weights, then compute a weighted sum of \(V\) to obtain the attention output. This process enables modeling intricate cross-sequence dependencies under mask constraints.

\underline{(ii) Feedforward layer} (FF) is a position-wise component that acts on the output of masked cross-attention (denoted as \(\mathbf{H}^b\)) to model complex non-linear mappings. It is formally defined as:
\begin{align}
    \operatorname{FF}(\mathbf{H}^b) = \operatorname{Linear}_2\big(\operatorname{ReLU}\big(\operatorname{Linear}_1(\mathbf{H}^b)\big)\big),
    \label{eq:feedforward}
\end{align}
where \(\operatorname{Linear}_1 \in \mathbb{R}^{d_h \times d_{ff}}\) and \(\operatorname{Linear}_2 \in \mathbb{R}^{d_{ff} \times d_h}\) are successive learnable linear layers, and \(d_{ff}\) (typically \(4d_h\)) is the intermediate hidden dimension. A dropout layer (dropout rate = 0.1 in our implementation) is inserted after the second linear projection as a standard regularization strategy to mitigate overfitting. Notably, the FF layer preserves the input dimensionality \(d_h\), ensuring consistency in sequence processing within the block.

\underline{(iii) Layer normalization} (LN) is an indispensable component applied to the hidden state matrix \(\mathbf{H}^b\) at the start of each block to stabilize hidden state distributions and accelerate training. It normalizes features across the hidden dimension \(d_h\) for each position (instead of across the sequence dimension), formulated as:
\begin{align}
    \operatorname{LN}(\mathbf{H}^b) = \gamma \cdot \frac{\mathbf{H}^b - \mathbb{E}[\mathbf{H}^b]}{\sqrt{\operatorname{Var}[\mathbf{H}^b] + \epsilon}} + \beta,
    \label{eq:layernorm}
\end{align}
where \(\mathbb{E}[\mathbf{H}^b]\) and \(\operatorname{Var}[\mathbf{H}^b]\) are the mean and variance of the hidden state matrix \(\mathbf{H}^b\) computed along the hidden dimension \(d_h\), \(\epsilon = 10^{-5}\) is a small constant for numerical stability, and \(\gamma, \beta \in \mathbb{R}^{d_h}\) are learnable affine parameters (scale and shift). This strategy alleviates internal covariate shift and preserves meaningful feature information via learnable parameters.

Collectively, the three components collaborate to update the input sequences. Taking \(\mathbf{A}^{b+1}\) as an example, its computation first aggregates residual cross-attention outputs from interactions between \(\mathbf{A}^b\) and the other two sequences (\(\mathbf{S}^b, \mathbf{R}^b\)):
\begin{align}
\mathbf{A}_{\text{attn}}^b &= \mathbf{A}^b + \operatorname{Attn}\big(\operatorname{LN}(\mathbf{S}^b), \operatorname{LN}(\mathbf{A}^b), \operatorname{LN}(\mathbf{A}^b)\big) \nonumber \\
& \quad + \operatorname{Attn}\big(\operatorname{LN}(\mathbf{R}^b), \operatorname{LN}(\mathbf{A}^b), \operatorname{LN}(\mathbf{A}^b)\big),
\label{eq:A_b+1_attn}
\end{align}
where the residual connection (\(\mathbf{A}^b + \cdots\)) mitigates vanishing gradients. The aggregated output is then processed by LN and FF to generate \(\mathbf{A}^{b+1}\):
\begin{align}
\mathbf{A}^{b+1} = \operatorname{FF}\big(\operatorname{LN}(\mathbf{A}_{\text{attn}}^b)\big),
\label{eq:A_b+1_final}
\end{align}
where \(\operatorname{LN}(\cdot)\) follows Eq. (\ref{eq:layernorm}), \(\operatorname{Attn}(\cdot)\) follows Eq. (\ref{eq:attention}), and \(\operatorname{FF}(\cdot)\) follows Eq. (\ref{eq:feedforward}). The computation of \(\mathbf{S}^{b+1}\) and \(\mathbf{R}^{b+1}\) adheres to the same logic (cross-attention with the other two input sequences + LN + FF). All operations preserve the hidden dimension \(d_h\) and sequence length \(M\), ensuring seamless information flow across consecutive blocks (\(b = 1, \dots, B\)).

\subsection{Constraint-aware Loss}\label{sec:Loss}
The Constraint-aware Loss (CL) is designed to enable the model to adaptively perceive and comply with core business constraints (i.e., CPA cost constraint and budget consumption constraint) in advertising auto-bidding during training~\cite{edelman2007internet,aggarwal2009general}. Unlike vanilla loss functions that solely optimize for reward maximization, this loss function introduces a dynamic penalty term \(P\) to penalize constraint-violating predictions, thereby achieving the dual goals of optimal performance and constraint compliance.

\paragraph{Construction of Penalty Term.}
The penalty term \(P\) combines two sub-terms ($P_{\text{CPA}}$ for excessive CPA and $P_{\text{BC}}$ for rapid budget consumption) via multiplication to enforce joint constraint compliance. The detailed formulation is as follows:

First, we compute the actual cost per action (CPA) at the end of a trajectory (i.e., the last time step $T$ of the trajectory):
\begin{align}
CPA_T = \frac{\sum_{i=1}^{I_T} x_i c_i}{\sum_{i=1}^{I_T} x_i v_i},
\label{eq:cpa_calc}
\end{align}
where $I_T$ denotes the total number of impression/click samples up to the last time step $T$ of a trajectory, $x_i$ is the delivery decision (e.g., bid or not) for the $i$-th sample, $c_i$ is the delivery cost (e.g., click cost) of the $i$-th sample, and $v_i$ is the number of valid conversions for the $i$-th sample. An indicator function is used to identify CPA constraint violations at the last time step $T$ of the trajectory:
\begin{align}
\mathbb{I}_{CPA_T > \theta} =
\begin{cases}
1, & \text{if } CPA_T > \theta \\
0, & \text{if } CPA_T \leq \theta
\end{cases},
\label{eq:cpa_indicator}
\end{align}
where $\theta$ is the advertiser-specified maximum acceptable CPA threshold (consistent with $C$ in Eq.~\eqref{eq:goal} for constraint consistency).

The CPA penalty term $P_{\text{CPA}}$ is activated only when the CPA at the last time step $T$ of the trajectory exceeds the threshold, with adaptive intensity:
\begin{align}
P_{\text{CPA}} =
\begin{cases}
\left(\frac{CPA_T}{C}\right)^{\alpha_1}, & \text{if } \mathbb{I}_{CPA_T > \theta} = 1 \\
1.0, & \text{otherwise}
\end{cases},
\label{eq:cpa_penalty}
\end{align}
where $C$ is the maximum allowable cost-rate threshold (Eq.~\eqref{eq:goal}) and $\alpha_1 > 1$ is the CPA penalty strength coefficient (larger $\alpha_1$ leads to heavier penalties for more severe violations; set to 2 in our experiments).

For the budget consumption constraint, we first calculate the budget consumption rate at the last time step $T$ of a trajectory:
\begin{align}
BC_T = \frac{\sum_{i=1}^{I_T} x_i c_i}{B},
\label{eq:bc_calc}
\end{align}
where $B$ is the total budget allocated to a single trajectory (Eq.~\eqref{eq:goal}). The budget consumption penalty term is defined as:
\begin{align}
P_{\text{BC}} = (BC_T)^{\alpha_2},
\label{eq:bc_penalty}
\end{align}
where $\alpha_2 > 1$ is the budget penalty coefficient (higher consumption rates result in more significant penalties; set to 2 in our experiments).

The total penalty term $P$ is the product of $P_{\text{CPA}}$ and $P_{\text{BC}}$, ensuring joint punishment for both constraints at the end of a trajectory:
\begin{align}
P = P_{\text{CPA}} \times P_{\text{BC}}.
\label{eq:total_penalty}
\end{align}
Notably, $P \geq 1.0$ for all scenarios: $P = 1.0$ (no penalty) only when the CPA constraint is satisfied and the budget consumption rate is 0 at the last time step $T$ of the trajectory; $P > 1.0$ if either constraint is violated or the budget is consumed rapidly, with larger $P$ indicating more severe violations.

\paragraph{Application to Action Loss and RTG Loss.}
The Constraint-aware Loss integrates the penalty term $P$ (denoted as $cpa\_p$ in code) into Action Loss ($\mathcal{L}_a$, bid prediction loss) and RTG Loss ($\mathcal{L}_r$, remaining reward prediction loss) via element-wise multiplication, enforcing constraint-aware weighting at the sample level.

The Action Loss (MSE loss for bid prediction) is defined as:
\begin{align}
\mathcal{L}_a = \frac{1}{N} \sum_{n=1}^N P_n \times (\hat{a}_n - a_n^*)^2,
\label{eq:action_loss}
\end{align}
where $N$ is the total number of samples, $P_n$ is the penalty term (derived from the last time step $T$ of the trajectory) for the $n$-th sample, $\hat{a}_n$ is the model-predicted bid parameter $\lambda_n$, and $a_n^*$ is the ground-truth target bid parameter. When the $n$-th sample's bidding strategy leads to constraint violations at the end of the trajectory, $P_n > 1.0$ amplifies the prediction error, forcing the model to prioritize correcting non-compliant bids.

The RTG Loss (MSE loss for remaining reward prediction) is formulated as:
\begin{align}
\mathcal{L}_r = \frac{1}{N} \sum_{n=1}^N P_n \times (\hat{r}_n - r_n^*)^2,
\label{eq:rtg_loss}
\end{align}
where $\hat{r}_n$ is the model-predicted remaining reward for the $n$-th sample, and $r_n^*$ is the ground-truth remaining reward. The penalty term $P_n$ (calculated at the last time step $T$ of the trajectory) ensures the model learns high-reward RTG distributions while complying with CPA and budget constraints for each sample by the end of the trajectory.

\paragraph{Total Loss.}
The final total loss function unifies the optimization of bid prediction and reward prediction under constraint awareness, defined as the sum of Action Loss and weighted RTG Loss:
\begin{align}
\mathcal{L}_{\text{total}} = \mathcal{L}_a + \lambda \cdot \mathcal{L}_r,
\label{eq:total_loss}
\end{align}
where $\lambda > 0$ is a hyperparameter that modulates the contribution of RTG Loss ($\mathcal{L}_r$); $\mathcal{L}_{\text{total}}$ balances bid accuracy and reward prediction quality, with both components constrained by the adaptive penalty term $P_n$ (derived from the last time step $T$ of the trajectory) to ensure adherence to advertising business rules by the end of each trajectory.
\section{Experiments}
\subsection{Experimental Setup}

\begin{table*}[t]
  \centering
  \caption{\textbf{Performance comparison}. Bold values are the highest scores, underlined values denote the best baseline results.}
  \vspace{-2mm}
  \setlength\tabcolsep{9pt}
  \scalebox{0.8}{
    \begin{tabular}{c|c|cccccccccc}
    \toprule
    \toprule
    Dataset & Budget & USCB  & CQL   & BCQ    & IQL   & CDT   & DT   & GAS   & GAVE  & C2  & \textit{Improve} \\
    \midrule
    \multirow{5}[2]{*}{AuctionNet} & 50\%  & 13.1  & 14.3  & 16.3  & 15.8  & 13.2  & 15.9  & 18.4  & \underline{19.6}  & \textbf{20.2} & 3.06\% \\
          & 75\%  & 16.2  & 17.5  & 23.1  & 24.7  & 20.0  & 22.3  & 27.5  & \underline{28.3}  & \textbf{28.5} & 0.71\% \\
          & 100\% & 19.5  & 24.6  & 30.3  & 31.5  & 32.1  & 33.3  & 36.1  & \underline{37.2}  & \textbf{38.4} & 3.23\% \\
          & 125\% & 25.8  & 28.4  & 35.2  & 36.4  & 34.3  & 36.2  & 40.0  & \underline{42.7}  & \textbf{43.0} & 0.70\% \\
          & 150\% & 33.7  & 34.8  & 36.9  & 40.4  & 38.6  & 39.0  & 46.5  & \underline{47.4}  & \textbf{48.4} & 2.11\% \\
    \bottomrule
    \bottomrule
    \end{tabular}%
    }
  \label{tab:overall}%
  \vspace{-4mm}
\end{table*}%

\paragraph{Dataset}
We conduct experiments on \textbf{AuctionNet}~\cite{su2024a}, a public large-scale real-world bidding dataset released by Alibaba and hosted on GitHub\footnote{https://github.com/alimama-tech/AuctionNet}. Unlike private non-open-source bidding logs, \textbf{AuctionNet} is derived from a manually built offline real advertising system, replicating the complexity of real-world ad auctions while ensuring public accessibility for research.

As the largest available public bidding dataset to our knowledge, it has a substantial scale: totaling over 500 million records with multiple advertising periods. Each period contains more than 500,000 impression opportunities, divided into 48 decision steps, and each opportunity involves bids from 50 agents. Each record includes core information such as predicted conversion value, bid amount, auction results, and impression outcomes.

\paragraph{Metrics}
Following previous studies~\cite{gao2025generative, li2024gas, su2024a, guo2024generative}, we conduct evaluations using varying budget ratios (50\%, 75\%, 100\%, 125\%, 150\%) from the original dataset to simulate real-world under-budget, full-budget, and over-budget scenarios. Performance is measured using the standard scoring metric for auto-bidding tasks:
\begin{equation}\label{equ:score1}
\begin{aligned}
penalty_j &= \min\left\{ \left( \frac{C_j}{\sum_i c_{ij} o_i / \sum_i p_{ij} o_i} \right)^\beta, 1 \right\}, \\
score &= \left( \sum_i o_i v_i \right) \times \min_{j=1}^J \{ penalty_j \},
\end{aligned}
\end{equation}
where we set $\beta=2$ following the standard configuration in prior works, $C_j$ is the constraint threshold for the $j$-th KPI, $c_{ij}$ and $p_{ij}$ are the cost and performance of the $i$-th sample under the $j$-th KPI, $o_i$ is the delivery outcome indicator, and $v_i$ is the sample valuation. Higher scores indicate better balance between performance and constraint compliance.

\subsection{Implementation Details}
All experiments are executed with a fixed batch size of 128 and a maximum of 10,000 training iterations to ensure sufficient convergence; model parameters are optimized using the AdamW optimizer~\cite{loshchilov2017decoupled} with a learning rate of $10^{-5}$, which is empirically determined to balance training stability and convergence speed, and specifically on the AuctionNet dataset, we set the hyperparameter $\lambda$ in Eq.~(\ref{eq:total_loss}) to 10 to appropriately weight the contribution of RTG Loss within the total loss function.

\subsection{Comparison with State-of-the-Arts}
To thoroughly validate the effectiveness of our proposed method C2 for advertising auction bid optimization, we conduct a comprehensive comparison between C2 and a diverse set of baseline methods (including USCB~\cite{he2021unified}, CQL~\cite{kumar2020conservative}, BCQ~\cite{fujimoto2019off}, CDT~\cite{liu2023constrained}, IQL~\cite{kostrikovoffline}, DT~\cite{chen2021decision}, GAS~\cite{li2024gas}, and the state-of-the-art baseline GAVE~\cite{gao2025generative}) on the AuctionNet dataset, covering five typical budget ratio settings (50\%, 75\%, 100\%, 125\%, and 150\%).

As illustrated in Table~\ref{tab:overall}, our C2 method achieves superior performance over all competing baselines across all budget configurations. Specifically, GAVE emerges as the best-performing baseline (marked by underlined values) in every budget scenario, yet our C2 consistently delivers the highest overall scores (highlighted in bold) across all tested budget ratios. Quantitatively, the performance improvement of C2 over GAVE ranges from 0.70\% (125\% budget) to 3.23\% (100\% budget), with an average gain of 1.96\% across all budget settings. Notably, the most substantial enhancement (3.23\%) is observed under the 100\% budget ratio—a standard operational scenario for advertising auctions—while C2 also maintains steady and meaningful gains in both under-budget (50\%, 75\%) and over-budget (125\%, 150\%) conditions.

These results suggest that our C2 method can effectively balance bid prediction accuracy and reward estimation reliability under budget constraints, contributing to consistent performance improvements over the state-of-the-art method GAVE across various budget scenarios on the AuctionNet dataset.



\subsection{Cross-correlations Analysis}
\begin{figure}[!t]
    \centering 
    \vspace{1mm} 
    \begin{subfigure}{1.0\linewidth}
        \centering
        \includegraphics[width=0.95\linewidth]{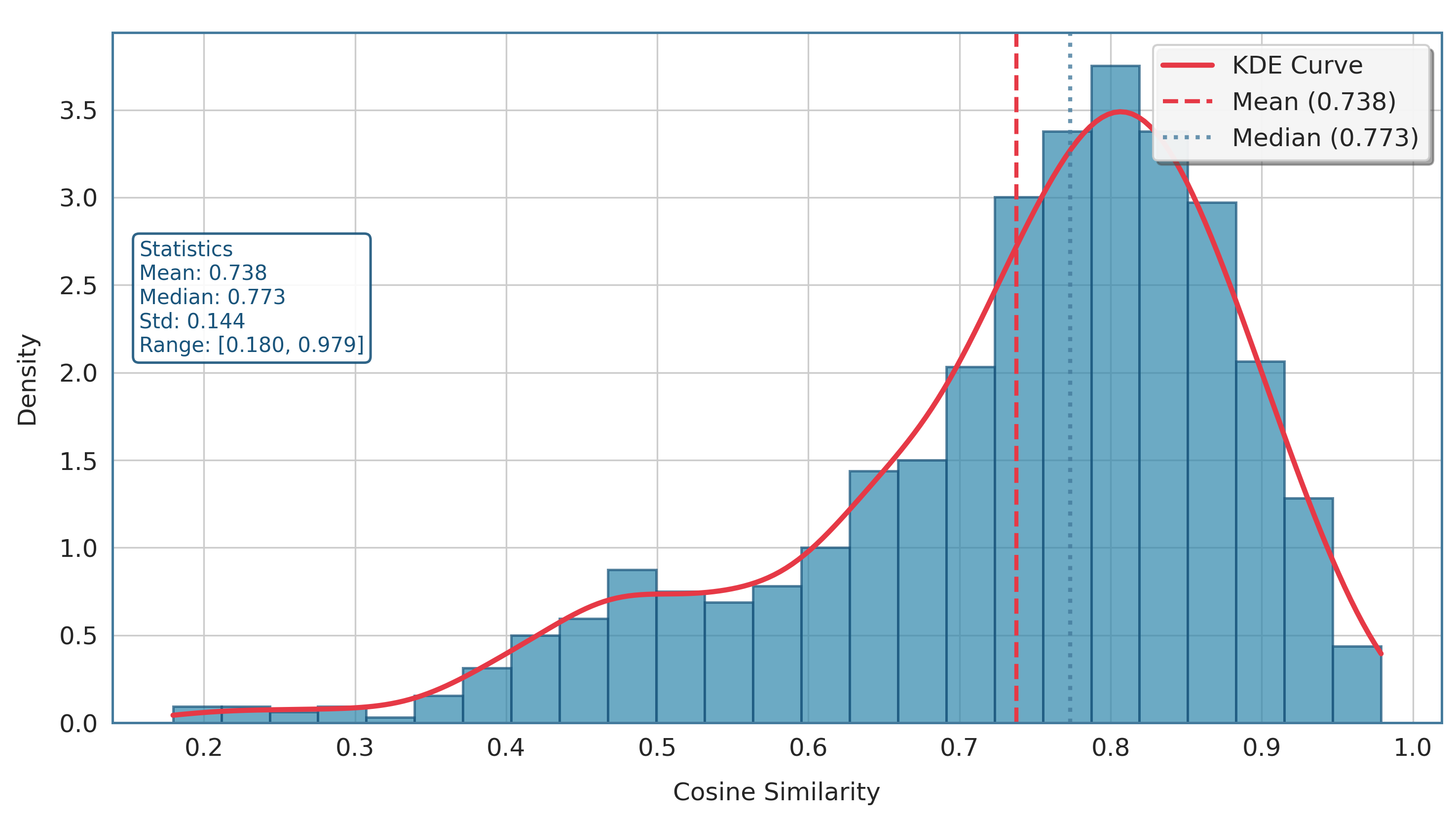}
        \caption{\textit{DT: Matched vs. shuffled embedding similarity (1000 samples)}}
        \label{subfig:vertical1}
    \end{subfigure}
    \vspace{0mm} 
    \begin{subfigure}{1.0\linewidth}
        \centering
        \includegraphics[width=0.95\linewidth]{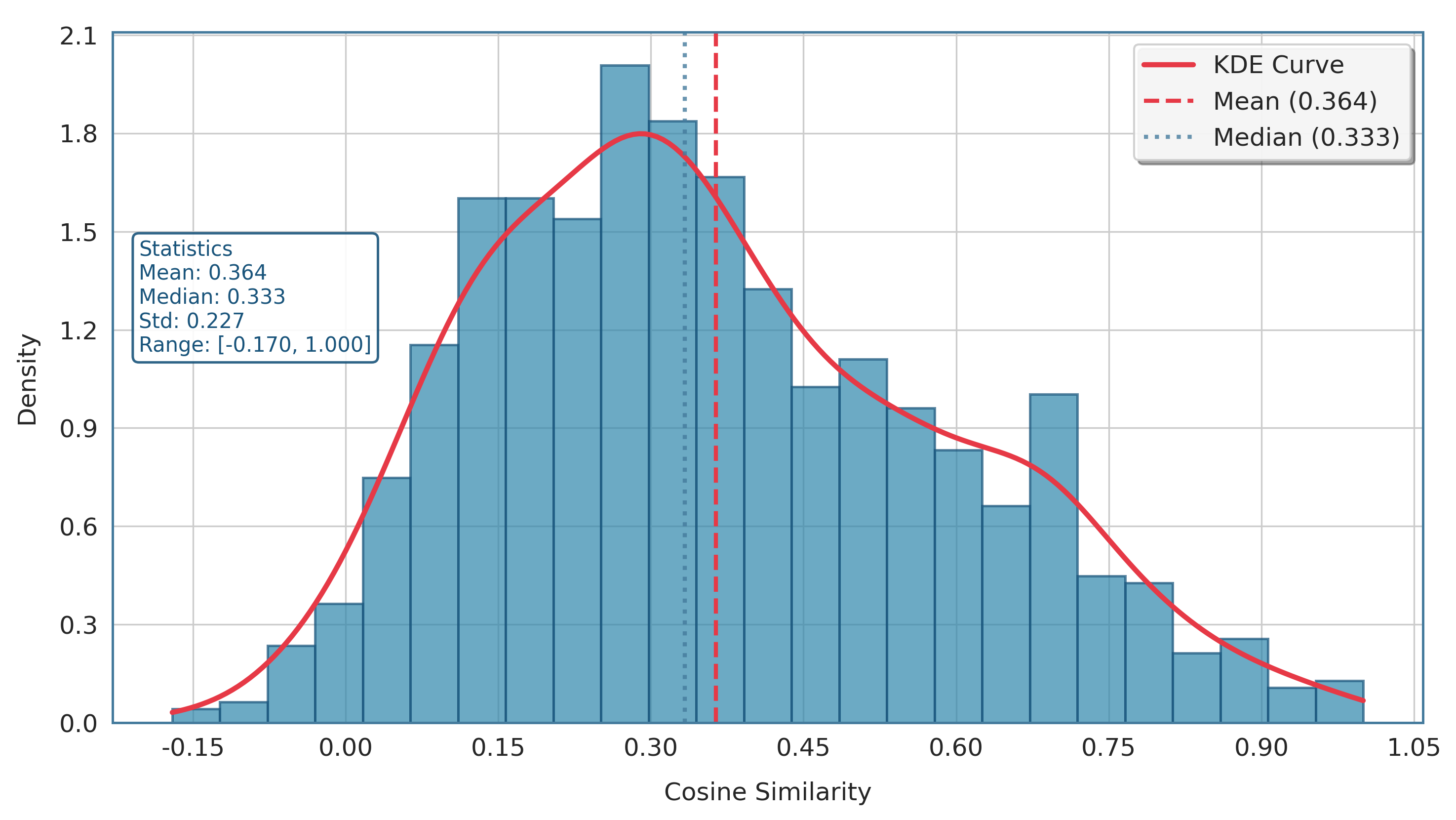}
        \caption{\textit{C2: Matched vs. shuffled embedding similarity (1000 samples)}}
        \label{subfig:cos}
    \end{subfigure}

    \vspace{-2mm}
    \caption{\textbf{Cross-correlation learning ability: C2 vs. DT}}
    \label{fig:vertical_results}
    \vspace{-5mm}
\end{figure}
To verify whether our C2 method alleviates the cross-correlation modeling deficiency of conventional Decision Transformer (DT) models (discussed in Section 1), we conduct a controlled experiment on the AuctionNet dataset. The core intuition is that a model with enhanced cross-correlation learning ability should generate significantly distinct embeddings for \textbf{matched} (original) and \textbf{shuffled} (state sequence randomly permuted) sequences—quantified by low cosine similarity between the two types of embeddings.

Specifically, we randomly sample 1000 instances from the AuctionNet dataset and extract two types of embeddings for each instance: (1) \textit{matched embeddings} from the first transformer block of the model when inputting the original sequential auction data; (2) \textit{shuffled embeddings} from the same block when inputting the sequence with its \textbf{state sequence randomly shuffled} (other components remain in original order). We compute the cosine similarity between matched and shuffled embeddings for all samples, with the distribution of similarity values shown in Figure~\ref{fig:vertical_results}.

Statistical results confirm that C2 effectively enhances cross-correlation modeling compared to DT:
\begin{itemize}
    \item For the vanilla DT model (Subfigure~\ref{subfig:vertical1}), the cosine similarity between matched and shuffled embeddings has a \textbf{high central tendency}: mean = 0.738, median = 0.773, Std = 0.144, and range = $[0.180, 0.979]$. High similarity values indicate DT's limited ability to distinguish between original and shuffled sequences, consistent with its insufficient cross-correlation modeling.
    \item For our C2 method (Subfigure~\ref{subfig:cos}), the cosine similarity distribution shows a \textbf{marked reduction in central tendency}: mean = 0.364 (50.7\% lower than DT), median = 0.333, Std = 0.227, and range = $[-0.170, 1.000]$. The substantially lower mean/median values prove C2 generates distinct embeddings for original and shuffled sequences, verifying that our method successfully mitigates DT's cross-correlation modeling issue.
\end{itemize}

These results validate that C2 effectively captures the interdependencies among sequential auction components, which is a key improvement over DT. By modeling cross-correlations in auction trajectories, C2 can leverage the temporal relational features of sequential data to make more optimal bid decisions in auto-bidding tasks.

\subsection{Ablation Study}

\begin{table}[!htb]
\renewcommand{\arraystretch}{1}
\renewcommand{\tabcolsep}{13pt}
\small
\caption{\textbf{Ablation study on the components of our proposed method}. (a) denotes the baseline model with only Decision Transformer (DT) component~\cite{chen2021decision}; (b) integrates Constraint-aware Loss (CL) into the DT baseline; (c) adopts the cross learning block-enhanced DT (CLB-DT) as the core backbone; (d) further incorporates CL into CLB-DT to maximize performance. Experiments are conducted under 100\% budget on AuctionNet.}
\vspace{-2mm}
\centering
\begin{tabular}{c|ccc|c}
    \toprule
    \toprule
    &\textbf{DT} & \textbf{CLB-DT} & \textbf{CL} &\textbf{Score}\\
    \midrule
    \textbf{(a)}&$\surd$& & & 33.3\\
    \textbf{(b)}&$\surd$& &$\surd$& 35.7\\
    \textbf{(c)}& &$\surd$& & 36.7\\
    \textbf{(d)}& &$\surd$&$\surd$&\textbf{38.4}\\
    \bottomrule
    \bottomrule
\end{tabular}
\label{table:ablation}
\end{table}
We perform an ablation study on the AuctionNet dataset under 100\% budget to evaluate the contributions of cross learning block-enhanced DT and Constraint-aware Loss in our C2 method (Table~\ref{table:ablation}).

The baseline DT model (a) achieves a score of 33.3. Adding CL to DT (b) improves the score to 35.7 (\textbf{7.2\%} gain), verifying CL's effectiveness in enforcing business constraints. Replacing DT with CLB-DT (c) yields a score of 36.7 (\textbf{10.2\%} gain over baseline), demonstrating CLB-DT's superiority in capturing auction trajectory correlations. Combining CLB-DT and CL (d) achieves the highest score of \textbf{38.4} (\textbf{15.3\%} gain), confirming the complementary synergy of the two components for optimal performance.

\section{Conclusion}
This paper proposes C2, a novel framework addressing two key limitations of Decision Transformer (DT) in auto-bidding: insufficient cross-correlation modeling among state-action-RTG sequences and indiscriminate behavior learning. C2 integrates two core components: Cross Learning Block (CLB) for enhanced inter-sequence dependencies via cross-attention, and Constraint-aware Loss (CL) for constraint-guided optimal trajectory learning.

Extensive offline evaluations on AuctionNet validate C2’s effectiveness, with up to 3.23\% gain over SOTA GAVE across diverse budget settings. Ablation studies confirm CL and CLB contribute 20.6\% and 24.0\% gains respectively over vanilla DT, achieving a 29.7\% combined improvement. Cross-correlation analysis further verifies CLB’s superiority in modeling sequence dependencies.

Future work will explore dynamic penalty adjustment for better adaptability and optimize the model structure to better adapt to auto-bidding scenarios.


\normalem
\bibliographystyle{ACM-Reference-Format}
\bibliography{bibfile}

\end{document}